\documentclass[conference]{IEEEtran}
\IEEEoverridecommandlockouts
\usepackage{cite}
\usepackage{amsmath,amssymb,amsfonts}
\usepackage{algorithmic}
\usepackage{graphicx}
\usepackage{textcomp}
\usepackage{xcolor}
\usepackage[colorlinks=true,linkcolor=blue]{hyperref}
\hypersetup{
 colorlinks=,
 citecolor=green,
 linkcolor=,
 urlcolor=}

\def\BibTeX{{\rm B\kern-.05em{\sc i\kern-.025em b}\kern-.08em
    T\kern-.1667em\lower.7ex\hbox{E}\kern-.125emX}}

\begin{document}

\title{Privacy-Preserving Remote Heart Rate Estimation from Facial Videos\\
\thanks{The authors would like to thank BMO Bank of Montreal and Mitacs for funding this research.}}

\author{\IEEEauthorblockN{Divij Gupta, Ali Etemad}
\IEEEauthorblockA{Dept. ECE \& Ingenuity Labs Research Institute \\
Queen's University, Canada\\
\{gupta.d,~ali.etemad\}@queensu.ca}}

\maketitle

\begin{abstract}
Remote Photoplethysmography (rPPG) is the process of estimating PPG from facial videos. While this approach benefits from contactless interaction, it is reliant on videos of faces, which often constitutes an important privacy concern. Recent research has revealed that deep learning techniques are vulnerable to attacks, which can result in significant data breaches making deep rPPG estimation even more sensitive. To address this issue, we propose a data perturbation method that involves extraction of certain areas of the face with less identity-related information, followed by pixel shuffling and blurring. Our experiments on two rPPG datasets (PURE and UBFC) show that our approach reduces the accuracy of facial recognition algorithms by over 60\%, with minimal impact on rPPG extraction. We also test our method on three facial recognition datasets (LFW, CALFW, and AgeDB), where our approach reduced performance by nearly 50\%. Our findings demonstrate the potential of our approach as an effective privacy-preserving solution for rPPG estimation.


\end{abstract}


\section{Introduction}
\label{sec:introduction}

Photoplethysmogram (PPG) is a non-invasive biosignal that measures blood volume changes in the vessels. PPG signals contain a wealth of medical information such as heart rate, blood pressure, heart rate variability, and others \cite{ppguse}. It also contains non-medical information such as emotion \cite{ppgemotion} and cognitive load \cite{ppgload}, among others. Compared to other biosignals PPG has the advantage of requiring simpler hardware and lower cost of operation, which has led to its wide usage for measuring important vitals \cite{ppgreview}. PPG is generally measured through contact sensors, which are often incorporated into devices such as wearables (e.g., smartwatches, smart bands, smart rings, earbuds, headphones, and others \cite{ppgreview,ppgear}). In this approach, infrared light is passed into the skin from a light-emitting diode (LED) and reflected back to a photodiode, both of which are placed on the surface of the skin. The varying amount of light reflected back from the blood vessels is tracked to comprise the signal. Recent advances in signal processing and machine learning have enabled a new and exciting way of acquiring PPG remotely in a non-contact way through cameras \cite{chrom,hrcnn}. By capturing and processing the light reflected from the surface of the skin \cite{verk}, the amount and speed of blood flow can be measured. This process is termed remote PPG (rPPG). 

Compared to other anatomical regions such as fingers, wrist, and legs, where standard PPG is often collected from, rPPG signals are stronger on the \textit{face} \cite{verk} given the volume of blood often flowing through it. Moreover, the face is the most readily accessible region to capture using a camera, hence is the pre-dominant body part for collecting rPPG data \cite{pure,ubfc}. To extract rPPG signal from facial videos, a large number of computer vision algorithms have been developed \cite{chrom,pos}, and advances in deep learning have further improved upon them \cite{deepphys,hrcnn}. However, the use and distribution of such datasets and algorithms is a deep concern in terms of \textit{privacy}. The use of facial data for rPPG estimation makes such applications highly sensitive as the face is one of the most crucial modes of biometrics which can be used to identify/authenticate and track individuals \cite{face_survey}. Recent research has also revealed that the security and integrity of deployed intelligent systems can be jeopardized which can lead to the leakage/reconstruction of sensitive information \cite{cafe, privacy_survey}.

To tackle this issue and to enable rPPG estimation from facial videos without risking or sacrificing privacy, we propose a simple yet highly effective pipeline for obtaining a privacy-preserving face representation to extract rPPG from. Our method first performs the extraction of pre-selected facial regions with the goal of excluding key identifying features, followed by shuffling of the pixels and blurring the outcome to obtain a privacy-preserving face representation. Through this, we destroy spatial consistency of facial regions at pixel level while maintaining the overall color intensity of the pixels which is crucial for the extraction of rPPG signals as demonstrated in \cite{chrom,pos}. We perform various experiments on two publicly available datasets, namely PURE \cite{pure} and UBFC \cite{ubfc}, and demonstrate that rPPG can be accurately measured using our face representations, while the detection of identities becomes excessively challenging.

Our contributions in this paper are threefold. (\textbf{1}) We propose a new face representation that preserves user identity and enhances privacy in the case of any leakage or reconstruction of data through malicious attacks. Our approach includes the selection of facial regions followed by the shuffling of pixels and blurring. (\textbf{2}) Our experiments on PURE and UBFC show that the proposed technique allows for rPPG to be effectively measured with minimal degradation while the recognition rate of identity is significantly reduced. In contrast, our comparison to other privacy-preserving representations for the face demonstrates that those techniques do not facilitate accurate estimation of rPPG. (\textbf{3}) Our thorough experiments validate the different design choices associated with the proposed method such as the order of shuffling, grouping of pixels, and others.

\section{Related Work}
In this section, we first present a literature review of prior work on rPPG estimation. We follow this by providing a short summary of existing works on facial recognition. Finally, we present a review of approaches for privacy-preserving solutions.

\subsection{Remote PPG}
Several classical methods have employed techniques such as color space transformations and signal processing to estimate rPPG. In CHROM \cite{chrom}, the mean color intensity of facial skin pixels was calculated and tracked for each frame to obtain three color intensity traces, one for each of the red, green, and blue (RGB) color channels. These traces were then bandpass filtered and linearly combined to derive the rPPG signal from the videos. In POS \cite{pos}, a similar approach to computing RGB traces was followed, but the traces were projected onto an orthogonal color space to estimate the rPPG signal. In 2SR \cite{2sr}, the authors utilized a slightly different technique where the skin pixels were identified, and a subspace of the skin pixels was created for each frame. The temporal rotation across these subspaces was then tracked to estimate the rPPG signals.

In recent years, deep learning techniques have further advanced rPPG estimation from facial videos. In \cite{hrcnn}, the authors employed a two-stage Convolutional Neural Network (CNN) model to estimate rPPG signals from face videos, which were then processed further to predict the heart rate. In \cite{phys}, the authors investigated various CNNs and Long Short-Term Memory-based spatial and temporal processing models for rPPG estimation. In \cite{deep}, a novel rPPG aggregating strategy was combined with a lightweight CNN architecture to adaptively combine rPPG signals from diverse skin regions. In \cite{deepphys}, a two-stream CNN model was proposed wherein the current frame (appearance) and its normalized difference with the next frame (motion) were processed in separate pathways. The network used intermediate fusions between the pathways to focus on the motion stream based on the appearance to extract rPPG. Also, some prior works have combined classical and deep learning approaches such as \cite{pulsegan}, wherein the authors first used \cite{chrom} to extract the rPPG signals and then refined them using a conditional General Adversarial Network (GAN) \cite{cgan}. Another line of work \cite{divij_aaai} has explored the use of self-supervised learning in rPPG to learn better features and reduce reliance on large amounts of labeled data for effective training.
 
\subsection{Face Recognition}
\label{related_prv}

Facial recognition is a widely studied problem in computer vision, for which various classical machine learning and deep learning solutions have been proposed. Classical facial recognition approaches use feature descriptors such as Local Binary Patterns \cite{lbp}, Scale-Invariant Feature Transform \cite{sift}, and others to generate feature descriptions of faces, which are then used for identification or verification. Similar to the classical approaches, deep learning methods use standard CNN architectures to generate high-dimensional feature embeddings from facial images, followed by classification \cite{face_survey}. Novel loss functions such as SphereFace \cite{sphere}, CosFace \cite{cos}, ArcFace \cite{arc}, and others, have been proposed to generate better separable embeddings in this context. Another avenue of research in face recognition focuses on using modalities such as depth-maps \cite{hardik1, hardik2}, light-field images \cite{light1,light2}, and others.

\subsection{Privacy Preservation}

Existing privacy-preserving approaches can be broadly divided into two categories depending upon the nature of their mechanism, into \textit{encryption} or \textit{perturbation}. Methods using encryption include Homomorphic Encryption \cite{homoenc}, Secure Multiparty Computation \cite{smc}, and others. However, encryption methods often incur high computation costs making them unsuitable for many deep learning systems. The other set of methods use perturbation in the training process in an effort to reduce the risk of successful privacy attacks. The most common perturbation technique used is Differential Privacy (DP) \cite{dpdl}. In DP, noise is added to the optimization process to prevent the model from strongly learning from any particular training sample. Specific to visual data, InstaHide \cite{insta} is a data perturbation method wherein every training sample is encoded using a weighted sum of itself and other training samples, after which the signs of the pixels of the composite image are randomly flipped. Another general privacy-preserving strategy for images has been explored in the works of \cite{le,ele}, wherein the authors divided a given image into blocks and then performed pixel perturbations within the blocks. 

Specific to datasets containing faces, privacy-preserving approaches include the addition of noise, masking, and blurring, among others \cite{gauss}. However, many of these approaches are known to be reversible \cite{deargan}, resulting in a need for privacy-preserving approaches for facial images with strong security and irreversibility. Other methods transform the face image to another domain such as in \cite{bdct_aaai}, wherein the authors take the Block Discrete Cosine Transform (BDCT) of the image, followed by channel-wise shuffling and combining to obtain the transformed image. Some methods seek to combine the concept of DP along with other transformations such as in \cite{bdct_eccv,peep} where the authors add noise to the BDCT and the Eigenface representation of the face image respectively to make them privacy-preserving. However, the effect of any privacy-preserving mechanism for rPPG extraction remains largely unexplored.

\begin{figure*}[t]
    \centering
    \includegraphics[width=1.6\columnwidth]{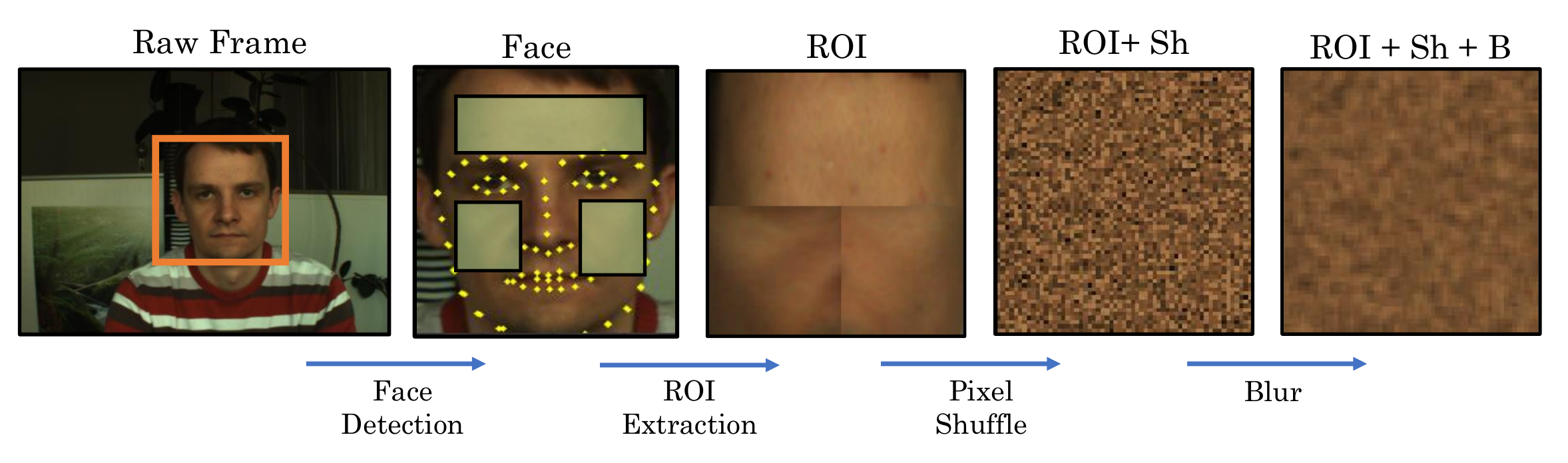}
    \caption{An overview of our proposed privacy preserving data perturbation pipeline.}
    \label{fig:process}
\end{figure*}

\section{Method}
\subsection{Problem Setup} 
Let $D$ be a dataset comprising face videos $F$ and corresponding PPG signals $P$ for $N$ subjects. Our goal is to design $Q$, a transformation with strong security, such that $M_{\theta'}(Q(F))\approx M_{\theta}(F) \approx P$, where $M$ is a deep learning model with learned parameters $\theta$, such that $Q(.)$ preserves the identities of the subjects in $D$, while $M$ is able to appropriately estimate the rPPG signal from $F$ or $Q(F)$.

\subsection{Method}
Let us assume a sample video $F_i$ comprising a total of $t$ frames $f^1, f^2, .... f^t$. These frames contain the face along with the background which serves no purpose in the rPPG estimation and might rather hinder it. Hence, we first detect the face in each frame using the Multi-Task Cascaded Convolutional Network (MTCNN) face detector \cite{mtcnn} and subsequently align and crop it. Next, we use Dlib \cite{dlib} to detect facial landmarks which are subsequently used to crop the left cheek, right cheek, and forehead from each frame. We crop these regions and use them instead of the entire face as the rPPG signals are stronger in these regions \cite{ppgro}. Moreover, we hypothesize that these regions contain less identity-related information, and are thus more suitable for privacy preservation. We crop the left and right cheeks first and then downsize the one with more height through interpolation such that the height of the two cheek regions becomes the same. After the resizing, we concatenate the cheek regions horizontally to form the whole cheek region. Next, we crop the forehead region and concatenate it (following resizing using interpolation) vertically with the cheek region to form the final RoI. Accordingly, we obtain frames $f^1_{RoI}, f^2_{RoI}, .... f^t_{RoI}$, which we then resize to a consistent size of $H \times W$ pixels, where we set $H$ and $W$ to 64.

Next, we apply a windowing operation with a window length of $T=128$, and a stride length of 8, to obtain smaller video clips which will be used for training and testing the approach. The same operation is applied to the corresponding PPG signals, $P$, to obtain $(V,S)$ training samples, where $V$ are the input clips and $S$ are the output corresponding PPG signals. $V$ comprises frames $f^1_{RoI}, f^2_{RoI}, ... f^{128}_{RoI}$, where $f^{i}_{RoI} \in \mathbb{R}^{64\times64\times3}$. We then flatten each frame into a $4096 \times 3$ array comprising ordered pixels. Next, we shuffle the ordering of these pixels randomly and reshape the array to obtain $f^i_{RoI+Sh}$ with dimensions $64 \times 64 \times 3$. Next, we blur $f^i_{RoI+Sh}$ by convolving $f^i_{RoI+Sh}$ with a $3 \times 3$ Gaussian kernel to obtain $f^i_{RoI+Sh+B}$. We perform this as we hypothesize that blurring helps in smoothing the shuffled pixels for relatively easier processing for rPPG extraction, while also causing loss of information for recovery of the original image, thereby providing a two-fold advantage for our use-case. We ensure that a set random order (key) is used for each particular sample while shuffling to maintain the pixel location coherence across all the frames of that particular sample. The process of RoI extraction from the face followed by shuffling and blurring comprises $Q$ (our privacy-preserving transform) and illustrated in Figure \ref{fig:process}. Here, we note that the search space of the key is $4096!$ which results in the strong security of our method.

\begin{table}[t]
\centering
\caption{Architectural details of 3D CNN used for rPPG estimation.} 
\scriptsize
\begin{tabular}{c c c c} 
\hline
\textbf{Block} &\textbf{Layers} & \textbf{Kernel~Size}      & \textbf{Output Size}        \\ \hline\hline
Input & - & - &  128$\times$64$\times$64$\times$3\\ \hline 
 ConvBlock1      &  \begin{tabular}[c]{@{}c@{}} Conv \\ AvgPool \end{tabular} &  \begin{tabular}[c]{@{}c@{}} 16, [1,5,5] \\ {[}1,2,2]  \end{tabular}  & \begin{tabular}[c] {@{}c@{}}128$\times$62$\times$62$\times$16 \\ 128$\times$31$\times$31$\times$16 \end{tabular}    \\  \hline 

 ConvBlock2      & \begin{tabular}[c]{@{}c@{}}Conv \\ AvgPool \end{tabular}     & \begin{tabular}[c]{@{}c@{}} 32, [3,3,3] $\times$2 \\ {[}1,2,2] \end{tabular}  & \begin{tabular}[c]{@{}c@{}}128$\times$31$\times$31$\times$32\\128$\times$15$\times$15$\times$32\end{tabular}    \\  \hline 

 ConvBlock3      & \begin{tabular}[c]{@{}c@{}}Conv \\ AvgPool \end{tabular}     & \begin{tabular}[c]{@{}c@{}}  64, [3,3,3] $\times$2 \\ {[}1,2,2] \end{tabular}  & \begin{tabular}[c]{@{}c@{}}128$\times$15$\times$15$\times$64\\128$\times$7$\times$7$\times$64\end{tabular}    \\  \hline 

 ConvBlock4    & \begin{tabular}[c]{@{}c@{}}Conv \\ GlobalAvgPool \\ Squeeze \\ Aggregation \end{tabular}     & \begin{tabular}[c]{@{}c@{}}  64, [3,3,3] \}$\times$3 \\ {[}1,7,7] \\ - \\ 1, [1]  \end{tabular}  & \begin{tabular}[c]{@{}c@{}}128$\times$7$\times$7$\times$64\\ 128$\times$1$\times$1$\times$64\\ 128$\times$64 \\ 128$\times$1 \end{tabular}    \\  \hline 
 Output & - & - &  128$\times$1\\ \hline 

\end{tabular}
\label{config}
\end{table}

Finally, we train a deep learning model to estimate the rPPG signal $S_r$ such that $S_r \approx S$ from the shuffled and blurred RoI clips. To this end, we use a 3D CNN as our model $M$, with learnable weights $\theta$. The model consists of 4 distinct convolutional blocks. The first block uses $1 \times 5 \times 5$ convolutional filters that extract spatial information from each frame of the video clips. The following three blocks use $3 \times 3 \times 3$ filters. Each convolution layer is followed by a ReLU activation and batch normalization. The detailed architecture of the model is presented in Table \ref{config}. To optimize $\theta$, we use the smooth L1 loss \cite{fastrcnn}. By combining the L1 and L2 losses, this loss enables switching between the two depending on the disparity between the amplitude values of the estimated rPPG signal $S_r$, and the ground-truth PPG signal $S$ allowing for smoother and more effective learning of weights. The loss is given by:
\begin{equation}
\small
\mathcal{L}(S_r, S) =
    \begin{cases}
      \frac{1}{2} \frac{(S_r - S)^2}{\beta}, ~~ |S_r - S| < \beta \\
      |S_r - S| - \frac{1}{2}*	\beta, ~~ otherwise .
    \end{cases} 
\end{equation}
where $\beta$ is a hyperparameter set to 0.3 for our method.

\section{Experiment Setup}

\subsection{Datasets}

\subsubsection{rPPG}
To test our method in terms of performance toward rPPG estimation, we experiment with two rPPG datasets, the descriptions of which are given below.

\noindent \textbf{PURE \cite{pure}.} This dataset comprises 60 facial videos and their corresponding PPG signals. There are a total of 10 subjects with each subject contributing 6 videos performing different movements such as steady sitting, talking, face rotation, and others. The PPG signals were collected using a Pulox CMS50E finger pulse oximeter at a sampling rate of 60 Hz while the videos were recorded using an Eco274CVGE camera at 30 frames per second. The videos have been saved in PNG format with a 640$\times$480 pixel resolution using lossless compression.

\noindent \textbf{UBFC \cite{ubfc}.} This dataset comprises 42 facial videos from 42 subjects while they were playing a time-sensitive mathematical game. The videos have been recorded with a Logitech C920 HD Pro webcam at 30 fps while the PPG signals have been recorded using a finger pulse oximeter Pulox CMS50E at 60 Hz sampling rate. The videos have been stored in uncompressed 8-bit format with a resolution of 640$\times$480 pixels.

\subsubsection{Facial Recognition}
To evaluate the ability of our method in reducing identification capability, we utilize a facial recognition system for benchmarking purposes. To this end, we also use four additional publicly available datasets \textbf{CASIA-Webface} \cite{casia}, \textbf{LFW} \cite{lfw}, \textbf{CALFW} \cite{calfw}, and \textbf{AgeDB} \cite{agedb} for facial recognition experiments which are summarized in Table \ref{tab:face_dat}.

\subsection{Evaluation Scheme and Metrics}
\noindent \textbf{RPPG.} For PURE, we use a train-test split of 6-4 subjects, while for UBFC, we use a 30-12 subject train-test split as done in previous works \cite{hrcnn, pulsegan}. Upon estimation of rPPG signals, heart rate is calculated using the Welch power spectrum method, similar to other works in the area \cite{hrcnn,deep}. We then take the average of the heart rate values to obtain an average heart rate for each test video and then compare it with the average ground-truth heart rate values to measure mean absolute error (MAE) and root mean square error (RMSE), both in beats per minute (bpm), along with correlation (R). \\
\noindent \textbf{Facial Identification.} We test the identifiability of the final perturbed images from both sets of datasets (rPPG datasets and facial recognition datasets). While the facial recognition datasets come with standard train-test protocols, this is not the case for the rPPG datasets. As a result, for PURE and UBFC, we randomly select 1000 facial frames for each subject from the dataset and apply the data perturbation scheme (RoI extraction, shuffling, and blurring) to generate the final perturbed images. Next, we use ArcFace \cite{arc}, a widely used face recognition algorithm to generate 512-dimensional embeddings of the images. We use, Principal Component Analysis (PCA) to transform the high-dimensional embeddings into lower-dimensional embeddings of 32 dimensions for easier classification. We then perform a 5-fold cross-validation using a Support Vector Machine (SVM) with a radial basis kernel. The final classification/identification (ID) accuracy serves as a utility measure for our privacy-preserving approach.

\subsection{Training}
\label{train}
For training the rPPG estimation model, we use a batch size of 8. We train the network for 15 epochs with 5e-4 as the learning rate for PURE, and 2e-4 for UBFC with Adam optimizer. For ArcFace, we train the model on CASIA-Webface for 25 epochs with a batch size of 180, and an initial learning rate of 1e-1 which is divided by 10 at 11, and 16 epochs. We use the SGD optimizer with a momentum of 0.9 and weight decay of 5e-4. All the codes were written in PyTorch and run on an NVIDIA Quadro RTX 8000 GPU.

\begin{table}[t]
\centering
\caption{Overview of the facial recognition datasets.} 
\scriptsize
\begin{tabular}{l |c |c} 
\hline
\textbf{Dataset} & \textbf{Identities} & \textbf{Total images}        \\ \hline
CASIA-Webface \cite{casia} & 10,575     &  494,414     \\ 
 LFW \cite{lfw}          & 5,749     &  13,233     \\ 
 CALFW \cite{calfw}   &   5,749      &  12,174     \\ 
 AgeDB \cite{agedb}  &   568   &       16,488\\ \hline 
\end{tabular}
\label{tab:face_dat}
\end{table}

\begin{table*}[t]
\centering
\caption{Comparison of various parameters for our proposed method on PURE and UBFC. 
}
\scriptsize
\setlength
\tabcolsep{10pt}
\begin{tabular}{l|c|c|c|c|c|c|c|c|c}
\hline
 & & \multicolumn{4}{c|}{\textbf{PURE}} & \multicolumn{4}{c}{\textbf{UBFC}} \\
\textbf{Input} & \textbf{Keys} & \textit{MAE}$\downarrow$ & \textit{RMSE}$\downarrow$ & \textit{R}$\uparrow$ & ID\% $\downarrow$ & \textit{MAE}$\downarrow$ & \textit{RMSE}$\downarrow$ & \textit{R}$\uparrow$ & ID\%$\downarrow$ \\ 
 \hline\hline
 Face  & - & 0.65 & 0.95 & 0.99 &  99.35  & 0.52  & 0.76 & 0.99 & 99.97  \\ 
 RoI  & - & \textbf{0.49 }& \textbf{0.78} & 0.99 & 98.99 & \textbf{0.44} & \textbf{0.65} & 0.99  & 99.93\\
    RoI+Sh & U & 1.23 & 1.71 & 0.99 &  60.20   & 0.79 & 1.14 & 0.99 & 43.49\\

 RoI+Sh+B & 1 & 1.23 & 1.68  & 0.99 &  98.79   & 0.61 & 0.71 & 0.99 & 99.86\\
 RoI+Sh+B  & 10 & 1.08 & 1.46 & 0.99 &  90.67 & 0.91 & 1.34 & 0.99 & 98.49 \\
 RoI+Sh+B  & 100 & 1.69 & 2.84 & 0.98 &  59.96  & 0.88 & 1.25 & 0.99 & 78.31 \\
  RoI+Sh+B  & 1000 & 1.20 & 1.98 & 0.99 & 46.81  & 1.01 & 1.52 & 0.99 & 42.08 \\
  RoI+Sh+B  & U & 0.96 & 1.30 & 0.99 & \textbf{46.19}  & 0.89 & 1.24 & 0.99 & \textbf{36.14} \\
 \hline
 RoI+Sh$_{2\times2}$+B  & 1 & 1.65 & 2.47  & 0.98 &  98.02   & 0.66 & 0.74 & 0.99 & 99.79\\
 RoI+Sh$_{2\times2}$+B  & 10 & 0.83 & 1.17 & 0.99 &  81.21 & 0.99 & 1.44 & 0.99 & 96.25 \\
 RoI+Sh$_{2\times2}$+B  & 100 & 1.11 & 1.63 & 0.99 &  48.98  & 0.65 & 0.81 & 0.99 & 55.53 \\
RoI+Sh$_{2\times2}$+B  & 1000 & 0.83 & 1.29 & 0.99 & 44.36  & 0.79 & 1.05 & 0.99 & 34.63 \\
RoI+Sh$_{2\times2}$+B  & U  & \textbf{0.79 }& \textbf{1.11} & 0.99 & \textbf{43.57}  & \textbf{0.53} & \textbf{0.69} & 0.99 & \textbf{30.80 }\\
 \hline 
 RoI+Sh$_{4\times4}$+B  & 1 & 1.45 & 2.35  & 0.99 &  97.91   & 0.96 & 1.36 & 0.99 & 99.81\\
 RoI+Sh$_{4\times4}$+B  & 10 & \textbf{0.65} & 0.78 & 0.99 &  83.39 & \textbf{0.66} &\textbf{ 0.89} & 0.99 & 97.54 \\
 RoI+Sh$_{4\times4}$+B  & 100 & 0.81 & 1.03 & 0.99 &  48.64  & 1.03 & 1.61 & 0.98 & 68.02 \\
  RoI+Sh$_{4\times4}$+B  & 1000 & 0.69 & \textbf{0.72} & 0.99 & 38.71  & 0.72 & 1.02 & 0.99 & 33.58 \\
  RoI+Sh$_{4\times4}$+B  & U  & 0.69 & 1.04 & 0.99 & \textbf{36.98}  & 0.69 & 1.09 & 0.99 & \textbf{27.17 }\\
 \hline
 RoI+Sh$_{8\times8}$+B  & 1 & 1.18 & 2.32  & 0.98 & 97.50   & 1.04 & 1.74 & 0.98 & 99.75\\
 RoI+Sh$_{8\times8}$+B  & 10 & 0.74 & 0.95 & 0.99 &  81.32 & 0.77 & 1.19 & 0.99 & 97.32 \\
 RoI+Sh$_{8\times8}$+B  & 100 & 0.80 & 1.15 & 0.99 &  45.67  & 0.81 & 1.28 & 0.99 & 70.78 \\
  RoI+Sh$_{8\times8}$+B  & 1000 &\textbf{0.69} & \textbf{0.72} & 0.99 & 32.67  & 0.63 & \textbf{0.79} & 0.99 & 29.10 \\
  RoI+Sh$_{8\times8}$+B  & U & 0.70 & 0.87 & 0.99 & \textbf{32.55}  & \textbf{0.61} & 0.85 & 0.99 & \textbf{20.83} \\
 \hline

\end{tabular}
\label{tab:all}
\end{table*}

\section{Results and Discussions}

\begin{figure}[t]
  \centering
  \includegraphics[width=0.6\columnwidth]{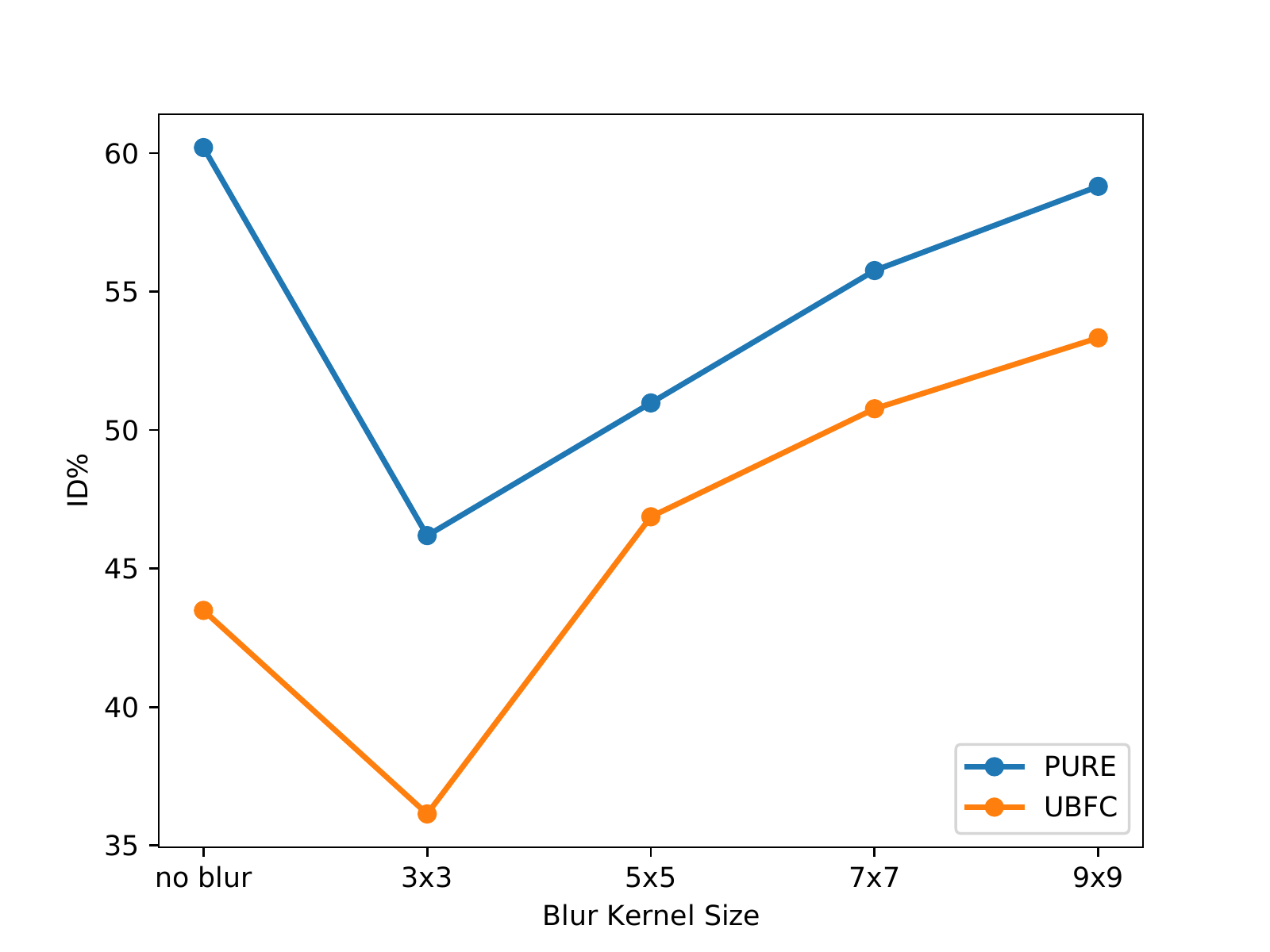}
  \caption{The impact of different kernel sizes used for blurring, on the identification accuracy, for both PURE and UBFC datasets.}
  \label{fig:recog_blur}
\end{figure}

\begin{figure}[t]
    \centering
    \includegraphics[width=1\columnwidth]{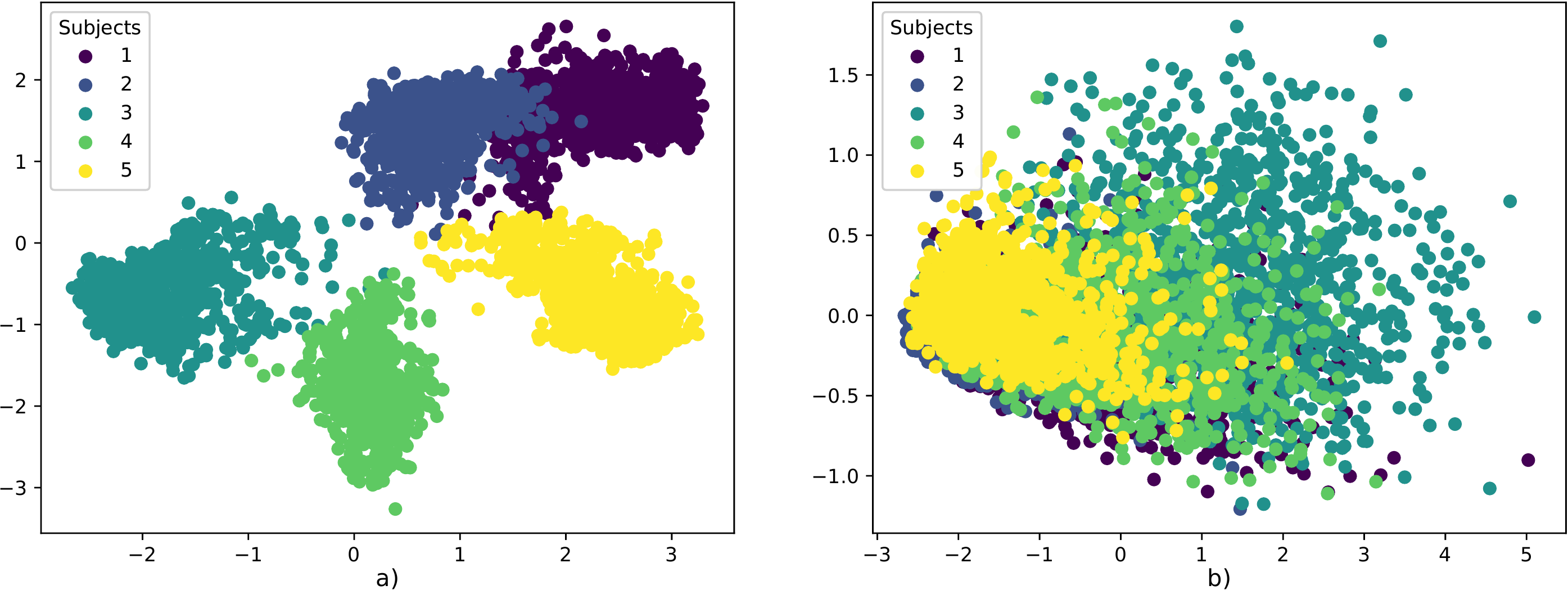}
    \caption{Visualization of ArcFace embeddings reduced to 2 dimensions with PCA for a) Face and b) RoI+Sh+B for 5 subjects in PURE.}
    \label{fig:emb}
\end{figure}

In Table \ref{tab:all}, we present the results of our experiments across all the metrics of rPPG estimation and facial identification. In the first part of the table, we observe that metrics for rPPG estimation (MAE, RMSE, R) improve when we use the cropped RoI as input for rPPG extraction, in comparison to the full face. We then compare the results of varying the number of choices for shuffling keys. We select the key from a choice of 1, 10, 100, and 1000 different possibilities and ultimately remove the bound of key choices (represented as \textit{U}), thereby randomizing the key selection to the entire available key space. We observe that introducing our data perturbation method causes some degradation across MAE and RMSE compared to the RoI-only setting. However, the performance remains within acceptable limits, especially for the \textit{U}-setting. In terms of identification, we observe that there is only a minimal reduction in accuracy when the RoI is used instead of the face. Moreover, there is little to no change in the accuracy when we introduce the shuffling and blurring step in the single key setting. As we increase the key choices, we observe some reduction in accuracy when the key choices are increased from 1 to 10, but a significant reduction in the accuracies upon increasing the choices from 10 to 100 and subsequently 1000, with the lowest accuracy being achieved in the \textit{U}-setting. 
The results of this experiment indicate that our approach significantly reduces the facial recognition performance, hence preserving privacy, while only resulting in a minor drop in performance for rPPG estimation. We also visualize the embeddings of ArcFace generated for the faces and the perturbed faces by our method in Figure \ref{fig:emb}. We clearly observe that our method disrupts the easily identifiable clusters of different subject classes. Finally, we experiment with the effect of blurring. First, as shown in the table, by comparing the performance of shuffling alone to shuffling plus blurring, we observe that in PURE, blurring has a positive impact on rPPG estimation as well as a reduction in identification accuracy, while in UBFC, there is a slight degradation in rPPG estimation, while still contributing to the reduction of identification capability. Moreover, when experimenting with different blurring window sizes, Figure \ref{fig:recog_blur} shows that 3$\times$3 is the optimal filter size for reducing identification accuracy.

\begin{figure}[t]
  \centering
  \includegraphics[width=0.5\textwidth]{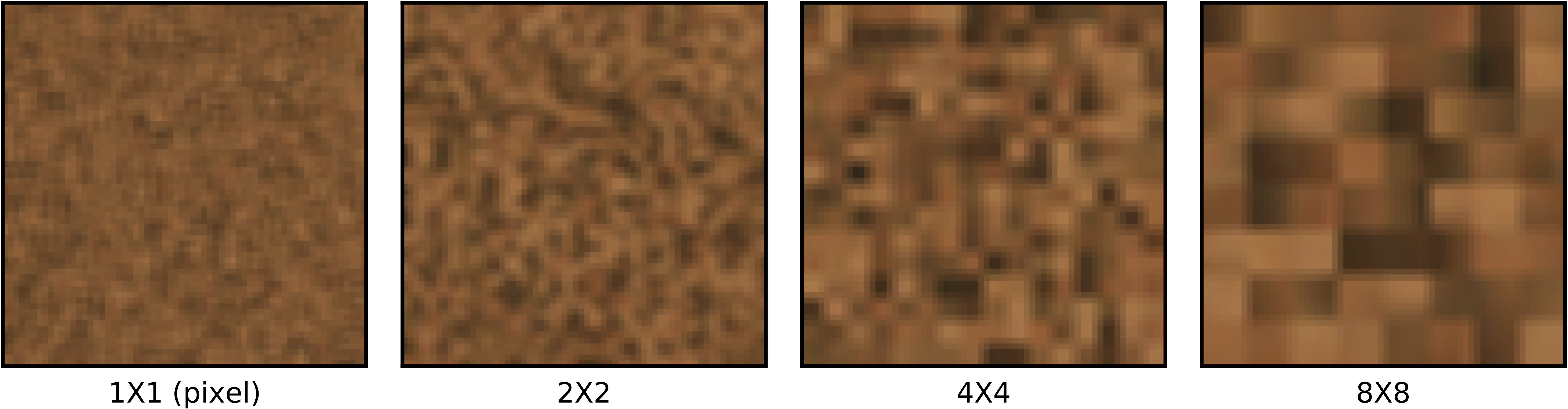}
  \caption{Visualization of pixel and patch-wise perturbed images.}
  \label{fig:patch}
\end{figure}

In the next parts of Table \ref{tab:all}, we also compare the effect of shuffling patches instead of pixels, wherein we first divide the frame $f^i_{RoI}$ into non-overlapping patches of size $P \times P$, flatten the image to obtain arrays of dimension $\frac{64^2}{P^2} \times P \times P \times 3$, and then shuffle the patches. Next, we reshape the outcome to obtain $f^i_{RoI+Sh_{P \times P}} \in \mathbb{R}^{64 \times 64 \times 3}$ and blur it. We observe a similar trend for the rPPG and ID metrics as observed in pixel-level shuffle settings where the rPPG metrics remained within acceptable limits as compared to the RoI-only setting, while the ID accuracies dropped considerably as the choice of shuffling keys were increased. However, we also observe that generally, the rPPG metrics improve and the identification accuracy further drops upon increasing the patch size. While this may suggest that rather than pixel-level shuffling, we should favour patch-wise shuffling, it is important to note that the grouping of pixels as patches greatly reduces the key space from $(64)^2!$ to $(64/P)^2!$ where $P \in \{2, 4, 8\}$. Moreover, since patches might be susceptible to Jigsaw-puzzle solver attacks \cite{jigsaw,etc}, we believe that pixel-level shuffling (\textit{U}-setting) may be a more secure approach. We visualize the effect of increasing patch sizes in our method in Figure \ref{fig:patch}. We can clearly observe that using patches can provide more information to the attacker and reduce the shuffle-key space.

\begin{table}[t]
\centering
\caption{Comparison with other privacy-preserving methods on rPPG estimation.}
\scriptsize
\setlength
\tabcolsep{3pt}
\begin{tabular}{l|c|c|c|c|c|c}
\hline
 & \multicolumn{3}{c|}{\textbf{PURE}} & \multicolumn{3}{c}{\textbf{UBFC}} \\
\textbf{Method} & \textit{MAE}$\downarrow$ & \textit{RMSE}$\downarrow$ & \textit{R}$\uparrow$ & \textit{MAE}$\downarrow$ & \textit{RMSE}$\downarrow$ & \textit{R} $\uparrow$ \\  
 \hline \hline 
 No perturbation  & 0.65 & 0.95 & 0.99  & 0.52  & 0.76 & 0.99 \\ 
  \hline
 BDCT \cite{bdct_aaai} & 14.44 & 15.50 & 0.10 & 13.54 & 16.50 & 0.03 \\ 
 Noise \cite{gauss} & 9.25 & 10.75 & 0.60 & 8.98 & 10.82 & 0.34 \\
 LE \cite{le} & 9.15 & 11.60 & 0.44 & 5.89 & 9.14 & 0.67\\ 
 InstaHide \cite{insta} & 2.33 & 3.02 & 0.97 & 2.77 & 4.02 & 0.92 \\
 Ours & \textbf{0.96} & \textbf{1.30} & \textbf{0.99} & \textbf{0.89} & \textbf{1.24} & \textbf{0.99} \\
 \hline
 
\end{tabular}
\label{tab:compare}
\end{table}

In Table \ref{tab:compare} we compare our method with several privacy-preserving methods \cite{bdct_aaai},  \cite{gauss}, \cite{le}, and \cite{insta} as described in Section \ref{related_prv}. 
First, we implement the Fast Face Image Masking utilizing BDCT proposed in \cite{bdct_aaai} to encode the facial images before feeding them into our rPPG  estimator. Next, we use a Gaussian Noise with a deviation of $\sigma^2 = 0.5$ similar to the baseline used in \cite{gauss}. We also use the encryption and adaptation strategy proposed in \cite{le} to modify the input face images. For these three methods, we use the same data perturbation for both training and inference. And finally, we implement InstaHide \cite{insta} with $k=2$. It is important to note that for InstaHide, during training, both the input as well as the output samples are mixed. However, since we do not want mixed/inseparable rPPG signals, we follow \cite{insta} and do not apply InstaHide during the inference process. The comparison results in Table \ref{tab:compare} demonstrate that our method results in better maintenance of rPPG features as the other privacy-preserving techniques result in significant drops in rPPG estimation performance.

\begin{figure}[t]
  \centering
  \includegraphics[width=0.45\textwidth]{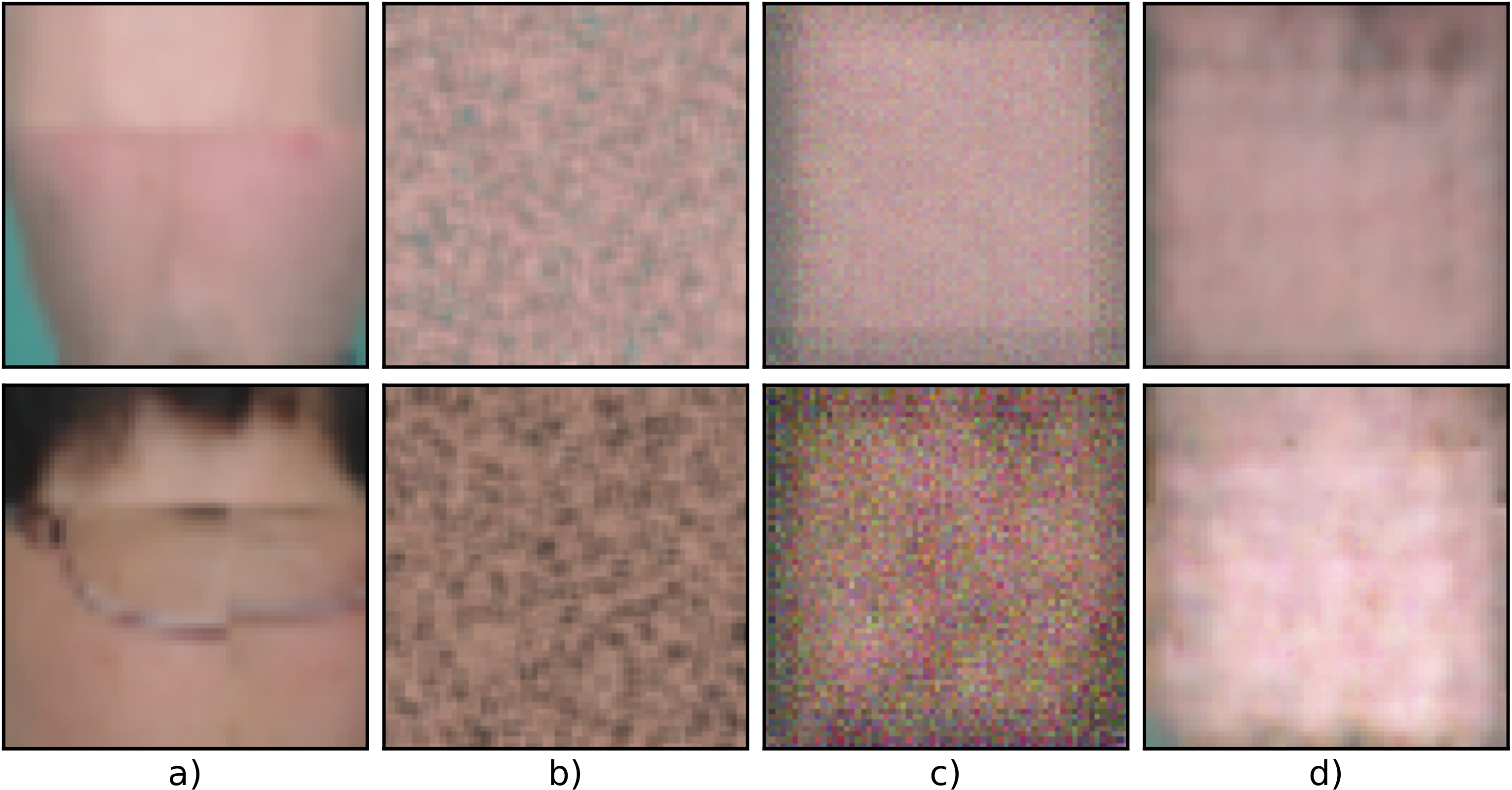}
  \caption{ Reconstruction attempts. a) RoI, b) RoI+Sh+B, c) Reconstructed RoI from UNet, and d) Reconstructed RoI from Pix2Pix.}
  \label{fig:ganop}
\end{figure}

To test the reversibility of the shuffling and blurring operation of our data perturbation scheme, we follow similar approaches to \cite{bdct_aaai,bdct_eccv} and implement a UNet \cite{unet} (batch size=512, learning rate=1e-1, epochs=20), and Pix2Pix GAN \cite{pix} (learning rate=2e-4, epochs=200) to try to learn a mapping between RoI and RoI+Sh+B images. We train the networks with pixel-shuffled and blurred RoI images as inputs and the original RoI images as outputs. In Figure \ref{fig:ganop}, we observe that recovering the RoI images is very difficult, further supporting our method.

\begin{table}[t]
\centering
\caption{Results of using our method on facial recognition datasets in terms of verification accuracy \%.} 
\scriptsize
\begin{tabular}{l| c| c| c| c} 
\hline
\textbf{Testing Input} & \textbf{Keys} &  \textbf{LFW} & \textbf{CALFW} & \textbf{AgeDB}         \\ \hline\hline
No perturbation & - & 99.35 & 93.03 & 93.53  \\ 
RoI & - & 65.78 &  55.65 &  51.78 \\ 
RoI+Sh & U  &  58.30 &  52.28 &  \textbf{49.91}  \\
RoI+Sh+B  & 1 & 57.25 &  53.33 & 50.11  \\ 
RoI+Sh+B  & 10 &  \textbf{53.76} &  52.58 &   50.93 \\ 
RoI+Sh+B  & 100 &  55.10 & 52.58 & 50.53   \\
RoI+Sh+B  & 1000 & 55.38 & 51.41 &  50.48   \\
RoI+Sh+B & U  & 54.16 &  \textbf{51.36} & 50.05   \\ \hline
\end{tabular}
\label{tab:arcface}
\end{table}

To further showcase the effect of our proposed data perturbation pipeline, we evaluate it on large-scale facial recognition tasks for which we use the ArcFace model trained on CASIA-Webface as described in Section \ref{train} and test it on LFW, CALFW, and AgeDB. The original datasets contain images of faces across varying poses, lighting, background, and even age. We test the ArcFace on the given face images, the extracted RoI, and the perturbed images. In Table \ref{tab:arcface}, we observe that using RoI instead of the face greatly reduces the verification accuracy, which further reduces upon the introduction of shuffling and blurring as part of our data perturbation scheme, and with increased randomness based on the key parameter for all the datasets.

\section{Conclusion and Future Work}
In this work, we proposed the detection of selected facial regions followed by pixel-shuffling and blurring as a means to preserve the identity of the subjects for rPPG extraction. Through comprehensive experiments, we validated the effect of different parameters of our proposed method on both the rPPG estimation as well as facial identification. We also showcased that our method causes a significant reduction in the performance of facial recognition systems when used for testing on public datasets. For future work, the effectiveness of our method could be explored in real-world scenarios, for instance on larger and more diverse datasets to evaluate its performance in different lighting conditions, camera angles, and facial expressions.

\bibliographystyle{myver}
\small
\bibliography{refs}

\end{document}